\pdfoutput=1
\documentclass[11pt]{article}

\usepackage[final]{acl}

\usepackage{times}
\usepackage{latexsym}

\usepackage[T1]{fontenc}

\usepackage[utf8]{inputenc}

\usepackage{microtype}

\usepackage{inconsolata}

\usepackage{graphicx}
\usepackage[breaklinks]{hyperref}
\usepackage{url}
\usepackage{booktabs}       
\usepackage{amsfonts}       
\usepackage{nicefrac}       
\usepackage{xcolor}         
\usepackage{paralist}
\usepackage{multirow}
\usepackage{subcaption}

\usepackage{amsmath}
\usepackage{amssymb}
\usepackage[export]{adjustbox}
\usepackage{stfloats}
\usepackage{tabularx}
\usepackage{makecell}
\usepackage{longtable}
\usepackage{xspace}
\usepackage{array}
\usepackage{caption}
\usepackage{svg}
\usepackage{mathtools}
\usepackage{parskip}
\usepackage{pgfplotstable}
\pgfplotsset{compat=1.18}
\pgfplotstableset{
    highlight bold/.style = {
        postproc cell content/.style={
            @cell content/.add={\boldmath}{},
        },
    },
}
\usepackage{siunitx}
\robustify\bfseries

\usepackage{tcolorbox}
\newcommand{\hlc}[2][yellow]{{%
    \colorlet{foo}{#1}%
    \sethlcolor{foo}\hl{#2}}%
}
\newcommand{\prompttype}[1]{\hlc[pink]{\textbf{\,#1\,}}}
\newcommand{\at}[1]{{\fontsize{10.5}{3}\selectfont\textit{@}}#1}

\newcommand{\shortsection}[1]{\vspace*{1ex}\noindent{\bf #1.}}
\newcommand{\shortsectionnp}[1]{\vspace*{.4ex}\noindent{\bf #1}}

\newcommand{\group}[1]{\mathcal{#1}}

\newcommand{\allocgap}[1]{\Delta\text{\small #1}}

\newcommand{\model}[1]{\textsc{#1}}

\usepackage[inline]{trackchanges}
\addeditor{YJ}

\usepackage{etoolbox}
\makeatletter
\patchcmd{\hyper@makecurrent}{%
    \ifx\Hy@param\Hy@chapterstring
        \let\Hy@param\Hy@chapapp
    \fi
}{%
    \iftoggle{inappendix}{
        \@checkappendixparam{chapter}%
        \@checkappendixparam{section}%
        \@checkappendixparam{subsection}%
        \@checkappendixparam{subsubsection}%
        \@checkappendixparam{paragraph}%
        \@checkappendixparam{subparagraph}%
    }{}%
}{}{\errmessage{failed to patch}}

\newcommand*{\@checkappendixparam}[1]{%
    \def\@checkappendixparamtmp{#1}%
    \ifx\Hy@param\@checkappendixparamtmp
        \let\Hy@param\Hy@appendixstring
    \fi
}
\makeatletter

\newtoggle{inappendix}
\togglefalse{inappendix}

\apptocmd{\appendix}{\toggletrue{inappendix}}{}{\errmessage{failed to patch}}

\title{Do Prevalent Bias Metrics Capture Allocational Harms from LLMs?}

\author{Hannah Cyberey, 
    Yangfeng Ji, David Evans \\
    Department of Computer Science\\
    University of Virginia\\
    Charlottesville, VA 22904\\
  \texttt{\{yc4dx,yangfeng,evans\}@virginia.edu} \\
}

\begin{document}
\maketitle

\begin{abstract}
\emph{Allocational harms} occur when resources or opportunities are unfairly withheld from specific groups. Many proposed bias measures ignore the discrepancy between \emph{predictions}, which are what the proposed methods consider, and \emph{decisions} that are made as a result of those predictions. Our work examines the reliability of current bias metrics in assessing allocational harms arising from predictions of large language models (LLMs). We evaluate their predictive validity and utility for model selection across ten LLMs and two allocation tasks. Our results reveal that commonly-used bias metrics based on average performance gap and distribution distance fail to reliably capture group disparities in allocation outcomes. Our work highlights the need to account for how model predictions are used in decisions, in particular in contexts where they are influenced by how limited resources are allocated.\footnote{Our code is available at: \url{https://github.com/hannahxchen/allocational-harm-eval}}
\end{abstract}
\section{Introduction}
The rise of large language models (LLMs) has raised concerns about potential harms in high-stakes decisions, such as lending~\citep{fu2021crowds}, hiring~\citep{bogen2018help}, and healthcare triage~\citep{rajkomar2018ensuring}. Recent orders in Europe~\citep{euaiact} and the U.S.~\citep{biden2023executive} have mandated audits to address AI risks including bias but left it unclear how to conduct effective audits.

Several works have conducted bias audits for LLMs in critical decision-making~\citep{tamkin2023evaluating,veldanda2023emily,haim2024s,armstrong2024silicone}. Yet, they focus on the \emph{predictions} models make, without considering how those predictions would be used to make decisions. Even when predictions appear to be unbiased, actual harms can arise from how they are used to make decisions~\citep{corbett2017algorithmic,mitchell2018prediction,kleinberg2018human}. As shown by \citet{dwork2018fairness}, evaluating models in isolation is insufficient to assert fairness without considering the context in which they will be deployed.

\textit{Allocational harms} arise if certain groups of people are deprived of access to resources or opportunities~\citep{crawford2017bias}. In settings where resources are limited and a model is used to prioritize options, there is a gap between \emph{predictions} and \emph{decisions}. It is unclear whether prevailing metrics, which measure bias in prediction outcomes, are sufficient to measure bias risks in applications where predictions are used for resource allocation.

\shortsection{Contributions}
To assess the potential harms of using LLMs for decision-making, we evaluate how well commmon bias metrics predict actual disparities in allocation outcomes. These metrics typically rely on average performance and distribution differences. We conduct this evaluation across ten LLMs on two allocation tasks (\autoref{sec:method}).  Our findings demonstrate that bias metrics based on predictions may not reliably reflect true disparities in outcomes (\autoref{sec:predictive-validity}). In addition, these metrics may sometimes identify models with greater disparities as less biased and exhibit inconsistent predictive abilities across different groups (\autoref{sec:metric-utility}). As a more reliable alternative, we propose the rank-biserial correlation, which demonstrates a strong correlation with actual allocation disparities.

\section{Background}
Algorithmic bias is commonly described as ``skew that produces a type of harm'' towards certain groups of people~\citep{crawford2017bias}. This can be further categorized into (i) \textit{harms of allocation}, which arise when models perpetuate an unfair distribution of resources (e.g., healthcare) or opportunities (e.g., jobs), and (ii) \textit{harms of representation}, which include stereotyping and misrepresentation.

\subsection{Measuring Bias} Proposed bias metrics are often formulated as the average group disparities in prediction outcomes based on established fairness definitions~\citep{czarnowska-etal-2021-quantifying}. The \textit{demographic parity gap} measures the difference in positive prediction rates between groups~\citep{agarwal2018reductions}. \textit{Equal opportunity} (EO), a relaxed notion of equalized odds, requires equal positive outcomes for qualified individuals~\citep{hardt2016equality}. The EO gap is thus the true positive rate differences between groups. For continuous predictions, group bias can be measured by the \textit{average score gap}~\citep{sicilia-alikhani-2023-learning}.
Several works consider the group distribution difference in prediction outcomes using distribution-based metrics such as Jensen–Shannon divergence~\citep{guo-etal-2022-auto}, Earth Mover's distance~\citep{huang-etal-2020-reducing}, and total variance distance~\citep{liang2022holistic}.

\subsection{Allocational Harms} 
\citet{blodgett-etal-2020-language} noted that NLP bias studies often lack clear and consistent motivations of what system behaviors are considered harmful and who is harmed and why. Out of thirty papers referencing allocational harms as motivation, they found only four actually propose measures or mitigations to address the harms~\citep{de2019bias,zhao-etal-2020-gender,romanov-etal-2019-whats,prost-etal-2019-debiasing}. Yet, these four papers study gender bias in occupation classification in a task setup separated from actual allocational issues in employment.

We find similar cases in subsequent works where the evaluation setups differ from allocation decision tasks in practice~\citep{kirk2021bias,lalor-etal-2022-benchmarking,shen-etal-2022-optimising,borchers-etal-2022-looking,van-aken-etal-2022-see}. Recent work has studied bias in LLMs used for hiring~\citep{veldanda2023emily,armstrong2024silicone,gaebler2024auditing} and other high-stakes decision scenarios~\citep{tamkin2023evaluating,haim2024s}. The evaluation methods adopted in these works only consider the average performance gap, measured from binary outputs or graded ratings. However, we show that this type of approach does not reliably reflect disparities in decision outcomes. We only find two closely related works that attempt to assess bias in resume ranking~\citep{bloomberg2024hiring,glazko2024identifying}. \citet{glazko2024identifying} evaluate disability bias in GPT-4 by the model's average preference difference between paired resumes. \citet{bloomberg2024hiring} inquires GPT-3.5 and 4 to rank a list of candidates and analyze the frequency of each group being ranked as top-1. We extend their work with more variations in resumes and conduct experiments on a wide range of open-weight LLMs.

\section{Method}
\label{sec:method}
We consider the allocation task as a top-$k$ ranking problem~\citep{cossock2006subset,clemenccon2007ranking}, where a fixed quota of $k\in \mathbb{N}$ candidates are selected among a pool of $n \!\gg\! k$ candidates. The goal is to determine a set of ``best'' candidates, with no particular emphasis on the relative order. We follow the LLM ranking method of \citet{zhuang-etal-2024-beyond} and rank the candidates in descending order of their prediction scores.

\subsection{Measuring Allocation Gaps}

Bias scores can be viewed as predictions of the allocation gaps in the following decision outcomes made with a model. An effective bias metric should yield a higher score for a group or a model when the outcome shows greater disparities. Given the decision outcomes of model $\mathcal{M}$ and allocation quota $k$, we measure allocation gaps using two common fairness criteria: demographic parity (DP)~\citep{agarwal2018reductions} and equal opportunity (EO)~\citep{hardt2016equality}.

The \emph{demographic parity gap} between group $\group{A}$ and $\group{B}$ is defined as:
\begin{equation*}
    \allocgap{DP}_{\mathcal{M}}(\group{A},\group{B}) = \phi_{\mathcal{M}}(\group{A}, k) - \phi_{\mathcal{M}}(\group{B}, k)
\end{equation*}
where ${\textstyle\phi_{\mathcal{M}}(\group{X}, k)}$ is the proportion of group $\group{X}$'s candidates selected.

We compute the \textit{equal opportunity gap} between group $\group{A}$ and $\group{B}$ as follows: 
\begin{equation*}
    \allocgap{EO}_{\mathcal{M}}(\group{A},\group{B}) = \psi_{\mathcal{M}}(\group{A}, k) - \psi_{\mathcal{M}}(\group{B}, k)
\end{equation*}
where $\psi_{\mathcal{M}}(\group{X}, k)$ is the rate of qualified candidates in group $\group{X}$ being selected.

\subsection{Bias Metrics}
\label{sec:metric-baselines}
Proposed bias metrics are often formulated as the average score or distribution difference between groups in prediction outcomes~\citep{czarnowska-etal-2021-quantifying,gallegos-etal-2024-bias}. 

\shortsectionnp{Average Performance Gap} computes the average score difference between group $\group{A}$ and $\group{B}$ as follows~\citep{sicilia-alikhani-2023-learning}:
$$\delta_{\mathcal{M}}(\group{A},\group{B}) = \frac{1}{|\group{A}|}\sum_{a\in\group{A}}s_a - \frac{1}{|\group{B}|} \sum_{b\in\group{B}}s_b$$
where $s_a$ is the prediction of candidate $a\in\group{A}$.

\shortsectionnp{Distribution-Based Metrics} measures score differences between groups using Jensen–Shannon Divergence (JSD)~\citep{lin1991divergence} and Earth Mover's Distance (EMD)~\citep{rubner1998metric}.

\shortsection{Rank-Biserial Correlation} We consider an alternative metric, \textit{rank-biserial correlation} (RB)~\citep{cureton1956rank}, which measures the correlation between group membership and ranking. It can be computed as the difference between the ratio of favorable pairs $f$ and unfavorable pairs $u$~\citep{kerby2014simple}:
\begin{align}
    RB_{\mathcal{M}}(\group{A},\group{B}) = f - u
\end{align}
where $f$ is the proportion of candidate pairs that model $\mathcal{M}$ prefers candidates from $\group{A}$ over $\group{B}$.

\subsection{Tasks}
\label{sec:task}
We evaluate settings where a model predicts the likelihood of a candidate match based on a description of an ideal candidate's qualifications. \autoref{app:experimental-setup}
provide further task details.

\shortsection{Resume Screening} Given a resume, the model evaluates a candidate's fit for a job position and outputs $\{\mathsf{No},\mathsf{Yes}\}$. We use four job positions from real job listings~\citep{bloomberg2024hiring}. We use GPT-3.5~\citep{chatgpt} to generate six resumes per position with varied hiring chances (high, medium, low), where high indicates qualified. Each candidate is represented by a first and last name on the resume. Each candidate pool includes one candidate sampled from each of the eight groups: $\{\text{Female}, \text{Male}\}\!\times\!\{\text{White}, \text{Black}, \text{Asian}, \text{Hispanic}\}$.

\shortsection{Essay Grading} The model is asked to rate each essay on a scale of $\left[1,5\right]$. We use the International Corpus Network of Asian Learners of English (ICNALE)~\citep{Ishikawa2013icnale}, which includes English essays written by second-language learners (L2) and first-language speakers (L1) on two topics. We consider qualified essays with a rating above average ($\geq$ the 50\textsuperscript{th} percentile)~\citep{ishikawa2024icnale}. Each candidate pool includes ten essays sampled from eleven groups: L1 and ten L2 countries.

\subsection{Experimental Setup}
We compute a bias score for each group $\group{A}\in \mathcal{G}\setminus \group{B}$ in comparison to a reference group $\group{B}$ (white males for resume screening and L1 speakers for essay grading). For each job position or essay topic, a total of $|\mathcal{G}|-1$ scores are produced for a model $\mathcal{M}$. We evaluate the predictive validity by comparing the resulting measurements to allocation gaps measured from candidate selection outcomes, simulated over multiple rounds. As JSD and EMD are non-directional, we compare them to the absolute value of $\allocgap{DP}$ and $\allocgap{EO}$. 

\shortsection{Models} We use ten LLMs with varied sizes and architectures: \model{Llama2 Chat} (7B, 13B)~\citep{touvron2023llama}, \model{Llama3 Instruct} (8B, 70B)~\citep{meta2024introducing}, \model{Gemma IT} (2B, 7B)~\citep{team2024gemma}, \model{StarlingLM 7B}~\citep{starling2023}, \model{StableLM Zephyr 3B}~\citep{StableLM-Zephyr-3B}, \model{StableLM2 Zephyr 1.6B}~\citep{bellagente2024stable}, and \model{TinyLlama Chat 1.1B}~\citep{zhang2024tinyllama}.
\begin{table}[tb]
    \begin{tabular*}{\linewidth}{@{}cc*{3}{S[table-format=-1.2{\textsuperscript{*}},table-space-text-post={*},detect-weight, detect-shape, detect-mode,]}@{}}
    \toprule
    & \multicolumn{2}{c}{Resume screening} & \multicolumn{2}{r}{Essay grading} \\
    Metric & $\allocgap{DP}$ & $\allocgap{EO}$ & $\allocgap{DP}$ & $\allocgap{EO}$ \\
    \midrule
    JSD & -0.19 & 0.48 & 0.79 & -0.19\textsuperscript{*} \\
    EMD & -0.09\textsuperscript{*} & -0.06\textsuperscript{*} & 0.86 & 0.48 \\
    $\delta$ & 0.13\textsuperscript{*}& -0.02\textsuperscript{*} & 0.89 & 0.70 \\
    RB & \bfseries 0.86 & \bfseries 0.88 & \bfseries 0.94 & \bfseries 0.89 \\
    \bottomrule
    \end{tabular*}
    \caption{Pearson correlation of bias metrics and allocation gaps. \textsuperscript{*} indicates p-value $>0.01$ with a 95\% confidence level.}
    \label{tab:metric-corr}
\end{table}
\begin{figure*}[tb]
\centering
\begin{subfigure}[b]{\linewidth}
\centering
\includegraphics[width=0.495\linewidth]{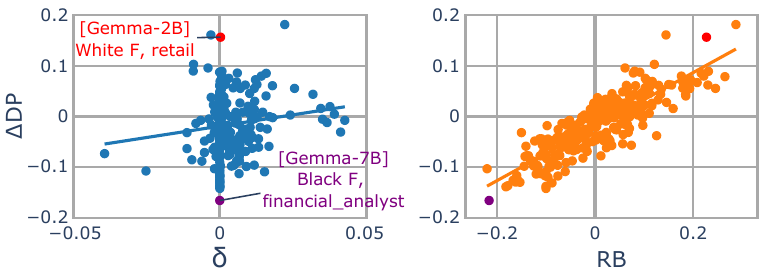}
\includegraphics[width=0.495\linewidth]{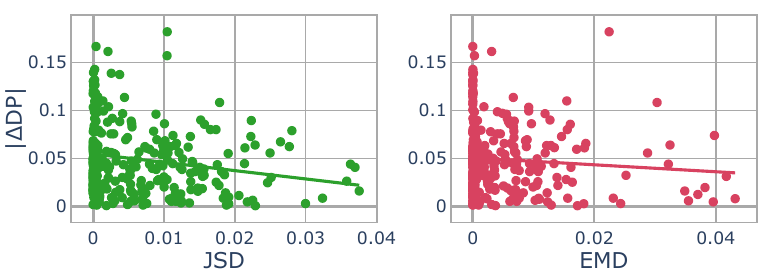}
\end{subfigure}
\caption{Measurement comparison between bias metrics and DP gap for resume screening, with $k=1$. Each point indicates a score measured for a group ${\textstyle \group{A}\in \mathcal{G}\setminus \group{B}}$, based on a model's predictions for a job position.}
\label{fig:overall-metric-corr-hiring}
\end{figure*}

\section{Results}
\label{sec:results}
This section shows results comparing bias metrics and allocation gaps in candidate selection outcomes based on LLM predictions. We first present the overall predictive validity, then the utility for model selection and informing bias risks.

\subsection{Predictive Validity}
\label{sec:predictive-validity}

\autoref{tab:metric-corr} reports the Pearson correlation of bias metric scores and allocation gaps for each task. It shows that $\delta$, JSD, and EMD do not predict allocational harms well. However, RB exhibits a strong correlation for both tasks, with a correlation $\geq 0.86$. EMD and $\delta$ show no correlation with $\allocgap{DP}$ and $\allocgap{EO}$ for the resume screening task. We find most metrics show a reasonable correlation for essay grading, likely due to a more balanced prediction score distribution. (see \autoref{sec:prediction-score-dist}).
 
\autoref{fig:overall-metric-corr-hiring} shows the data points for computing the correlations with $\allocgap{DP}$ for resume screening (second column in \autoref{tab:metric-corr}). Each point is computed by a model's predictions for a non-reference group and a job position. Many scores of $\delta$ exhibit close to zero bias with respect to white males, indicated by points along the y-axis where $\delta=0$. E.g., \model{Gemma IT 2B} for white females and the retail position. Yet, some of them show a larger allocation gap than ones with a higher $\delta$. 

\subsection{Metric Utility for Model Selection}
\label{sec:metric-utility}
When a metric is used in a model audit, it could be used to determine if a model meets some required threshold scores or decide between a set of candidate models. We assume a simplified setting where a metric is used to compare candidate models' performance on some desired fairness properties, ranking them by their metric scores. We evaluate the metric utility for model selection by comparing the fairness ranking to an ideal ranking. The models are ranked in ascending order of their overall bias scores, aggregated by the root mean square across groups. Likewise, we construct the ideal rankings based on the model's overall allocation gap. 

Suppose a bias metric produces a fairness ranking $\tau$, and the ideal ranking is $\sigma$. We compute
the normalized discounted cumulative gain (NDCG) following \citet{jarvelin2002cumulated} as:
\begin{align*}
    \mathsf{NDCG}\at{N}(\tau) &= \frac{\mathsf{DCG}\at{N}(\tau)}{\mathsf{DCG}\at{N}(\sigma)}
\end{align*}
where $N$ is the rank cutoff. $\mathsf{DCG}$ emphasizes the ``best'' ideal models and imposes a penalty when they are low-ranked. 

\begin{figure}[tb]
  \centering
  \includegraphics[width=\linewidth]{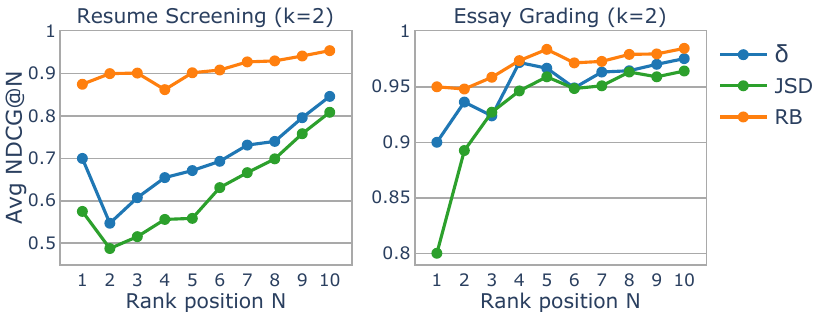}
 \caption{Average NDCG$\at{N}$ in ranking model fairness, comparing to ideal rankings based on $\allocgap{DP}$. EMD yields the same results as $\delta$.}
 \label{fig:model-rank-ndcg}
\end{figure}

\autoref{fig:model-rank-ndcg} reports the average NDCG based on fairness criteria $\allocgap{DP}$ with quota $k=2$ for each task. RB consistently performs better than other bias metrics with an average NDCG$\at{10}$ $\geq 0.95$ on both tasks. NDCG$\at{1}$ indicates how close the top-1 model is to the top of the ideal ranking.

\begin{figure}[tb]
  \centering
  \includegraphics[width=\linewidth]{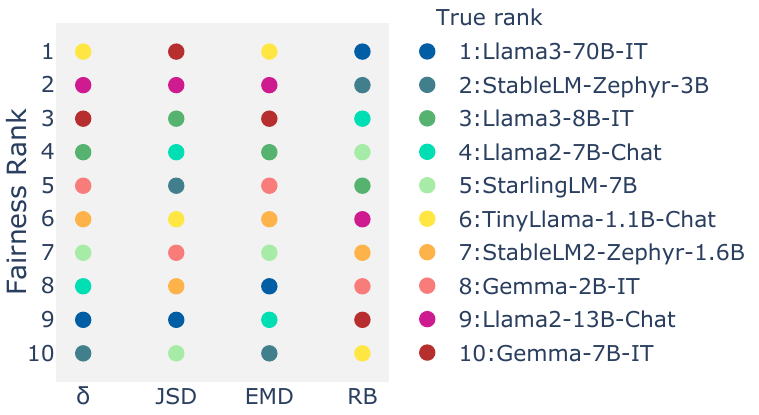}
  \caption{Model fairness ranking for the resume screening task with selection quota $k=2$. The true rank order is based on $\allocgap{DP}$. Existing bias metrics often rank more biased models as more ``fair''.}
 \label{fig:model-DP-rank}
\end{figure}

In \autoref{fig:model-DP-rank}, we further compare the fairness ranking of models among bias metrics for the resume screening task. The ranking of RB aligns more closely with the ranking based on $\allocgap{DP}$, whereas other bias metrics tend to rank more biased models higher. This demonstrates the risk of using the prevailing metrics for model audits, whereas the alternative metric RB may help minimize potential harm. We provide the ranking per job position in Appendix~\ref{app:metric-utility}.

\begin{figure}[tb]
  \centering
  \includegraphics[width=0.9\linewidth]{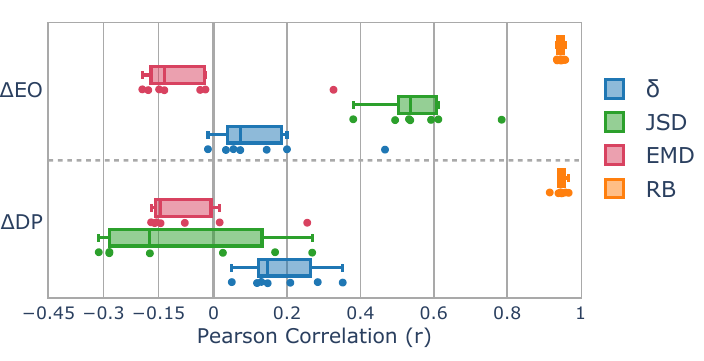}
  \caption{Bias metric and allocation gap correlation by group in resume screening with $k=2$. Common bias metrics exhibit varying correlations among groups.}
 \label{fig:metric-corr-by-group}
\end{figure}

\shortsection{Predicting bias across groups} \autoref{fig:metric-corr-by-group} shows the correlation of bias metric and allocation gap measured by group across all models. Distribution-based metrics and $\delta$ show significant variations in their ability to predict allocation gaps in resume screening outcomes. In some cases, they even show a positive correlation for some groups while exhibiting a negative correlation for the other groups. In contrast, RB exhibits consistent performance for different groups. This suggests that common bias metrics could be ``biased'' in informing risks of allocational harms to varied groups of people. 

\begin{figure}[tb]
    \centering
    \includegraphics[width=0.95\linewidth]{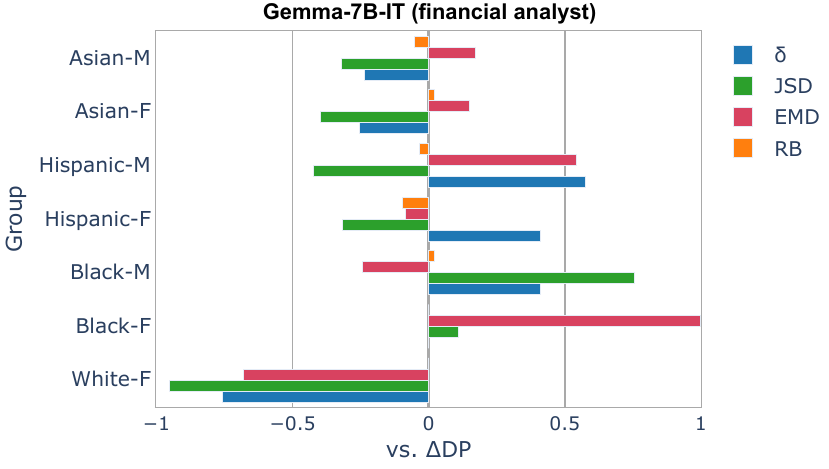}
    \caption{Difference between bias scores and $\allocgap{DP}$, after normalizing to $\left[0,1\right]$, across groups with $k=2$. A negative difference indicates $\allocgap{DP}$ is underestimated.}
    \label{fig:gemma7b-financial-analyst}
\end{figure}

To illustrate the impacts of using a metric, we measure the difference between the bias score and allocation gap for each non-reference group after normalizing the scores to $\left[0,1\right]$. In \autoref{fig:gemma7b-financial-analyst}, all metrics except RB underestimate the degree of negative impact on white females. The negative impact on Hispanic males is overestimated by $\delta$ and EMD but underestimated by JSD.

\subsection{Analysis}
\label{sec:prediction-score-dist}
\begin{figure}[tb]
  \centering
  \includegraphics[width=0.95\linewidth]{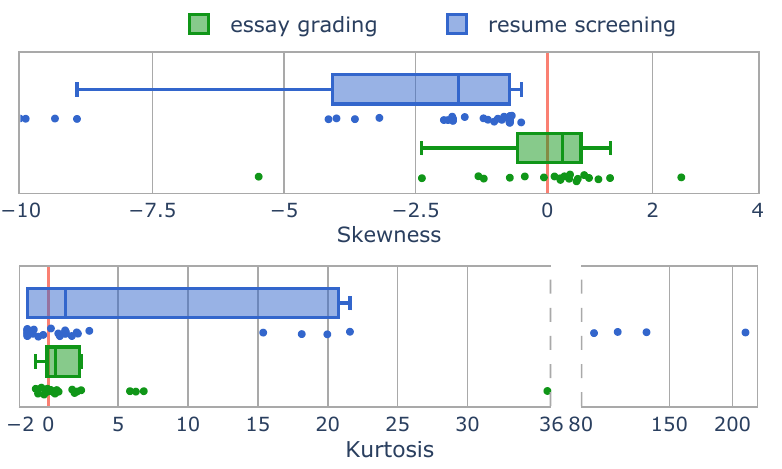}
  \caption{Skewness and kurtosis of all ten models' prediction score distribution per task. Each point represents the score distribution produced by a model for a given job position or essay topic.}
 \label{fig:pointwise-score-dist}
\end{figure}

\autoref{fig:pointwise-score-dist} depicts the skewness and kurtosis of the prediction score distributions produced by all ten models for both tasks. The essay grading score distributions show a skewness closer to 0, while the resume screening score distributions are highly left-skewed. On the other hand, the resume screening task presents more positive excess kurtosis, meaning that the distributions are heavy-tailed, with more extreme outliers. (A standard normal distribution has a kurtosis of 3.) This may explain why the traditional bias metrics show a better correlation with the allocation gaps on the essay grading task than the resume screening task. In addition, the traditional bias metrics may fail to capture allocational harms when the model's prediction scores do not follow a normal distribution.
\section{Discussion}
Our findings reveal that common bias metrics for evaluating LLMs do not capture allocational harm. While final decisions may depend on human decision-makers or other external factors, a reliable measurement is crucial for estimating the potential risks of a model. In fact, in settings of unfamiliar domains and objective tasks, humans tend to rely more on model predictions~\citep{yeomans2019making,chiang2021you,passi2022overreliance}. \citet{green2019disparate,green2021algorithmic} have shown that algorithmic risk assessments not only alter human decisions but exacerbate racial disparities.

The goal of an audit is to determine if it is acceptable to deploy a model. Although audits will always be imperfect since they require making predictions about how the model will behave on future data, it is essential that we develop methods for auditing models that reliably measure potential harms in the way models will be used in deployment. Our results demonstrate that metrics too far removed from how a model will be used may fail to adequately measure how well the model will perform as deployed.

\bibliography{anthology,custom}
\onecolumn
\appendix

\section{Experimental Setup}
\label{app:experimental-setup}

\subsection{Task Setup}

\shortsection{Resume Screening} We construct a dataset that includes instructions and resume templates based on descriptions of four real job positions (software engineer, HR specialist, financial analyst, and retail) used in Bloomberg's bias audit study~\citep{bloomberg2024hiring}. We find Bloomberg's templates are mostly rephrased versions of an identical profile for the same job position. Thus, we prompted GPT-3.5~\citep{chatgpt} to generate resume templates for each job description. Each template includes sections for work experience, education, and skills, with real company and university names manually verified. Each group is represented by 100 common first and last names based on data from the Social Security Administration and voter files in US~\citep{rosenman2023race}.

\shortsection{Essay Grading} ICNALE consists of 5.6K English essays written by 2.6K second language (L2) college students from 10 Asian countries and 200 first language (L1) speakers~\citep{Ishikawa2013icnale}. 140 essays include ratings (0$\sim$100) from L1 English speakers. Each writer was asked to write opinion essays on two topics: 
\begin{enumerate}
    \item \textbf{PTJ}: It is important for college students to have a part-time job.
    \item \textbf{SMK}: Smoking should be completely banned at all the restaurants in the country.
\end{enumerate}

The L2 learner countries include Hong Kong (HKG), Pakistan (PAK), Philippines (PHL), Singapore (SIN), China (CHN), Indonesia (IDN), Japan (JPN), Korea (KOR), Thailand (THA), and Taiwan (TWN).
\\
\begin{table}[!h]
\setlength\extrarowheight{8pt}
\small
    \centering
    \begin{tabular*}{\linewidth}{@{}ccccccc@{}}
    \toprule
    Task & \makecell{Prediction\\Outcome} & Groups ($\group{G}$) & Ref. group & Pool size & max $k$ & Rounds \\
    \midrule
    Resume Screening & \makecell{Good fit for\\job position} & \makecell{$\{\text{Female}, \text{Male}\} \times$\\$\!\{\text{White}, \text{Black}, \text{Asian}, \text{Hispanic}\}$} & White Male & 8 & 5 & 1800 \\
    Essay Grading & Essay's rating & \makecell{HKG, PAK, PHL, SIN, CHN, IDN,\\JPN, KOR, THA, TWN, ENS} & ENS & 10 & 5 & 1200 \\
    \bottomrule
    \end{tabular*}
    \caption{Parameters used for simulating candidate selection.}
    \label{tab:task_params}
\end{table}

\subsection{LLM Ranking}
\label{app:llm-ranking}
This section explains the method for computing the ranking scores.

Suppose $Y$ is a set of relevance labels, where each $y\in Y$ corresponds to a relevance value $\gamma_{y}$. Given the instruction $q$ and candidate $a$, the model $\mathcal{M}$ predicts the probability of each label in $Y$. The ranking score of candidate $a$ is defined as~\citep{zhuang-etal-2024-beyond}:
$$\mathit{score}_{q,\mathcal{M}}(a) = \sum_{y\in Y} P_n(\mathcal{M}_q(a), y)\cdot \gamma_{y}$$
where $P_n$ is the normalized output probability of $y$ over $Y$. The score is assumed to encode the relevance or fitness of candidate $a$. For the resume screening task, we consider $Y=\{\mathsf{No},\mathsf{Yes}\}$ with $\gamma_{y}\in\{0, 1\}$. For the essay grading task, the relevance labels and values are on a rating scale of $\left[1,5\right]$.

\newpage

\section{Additional Results}

\subsection{Predictive Validity}

\begin{figure}[h!]
\centering
\begin{subfigure}[b]{\linewidth}
\centering
\includegraphics[width=0.495\linewidth]{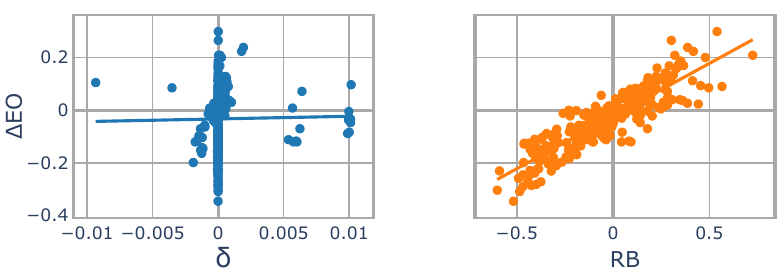}
\includegraphics[width=0.495\linewidth]{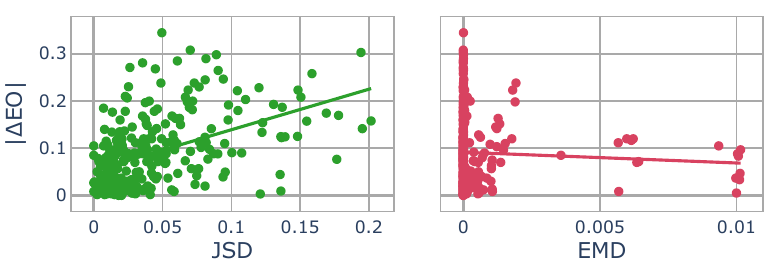}
\end{subfigure}
\caption{Bias metrics (x-axis) and allocation gaps (y-axis) for \textsc{resume screening}, with quota $k=1$.}
\end{figure}

\begin{figure}[h!]
\centering
\begin{subfigure}[b]{\linewidth}
\centering
\includegraphics[width=0.495\linewidth]{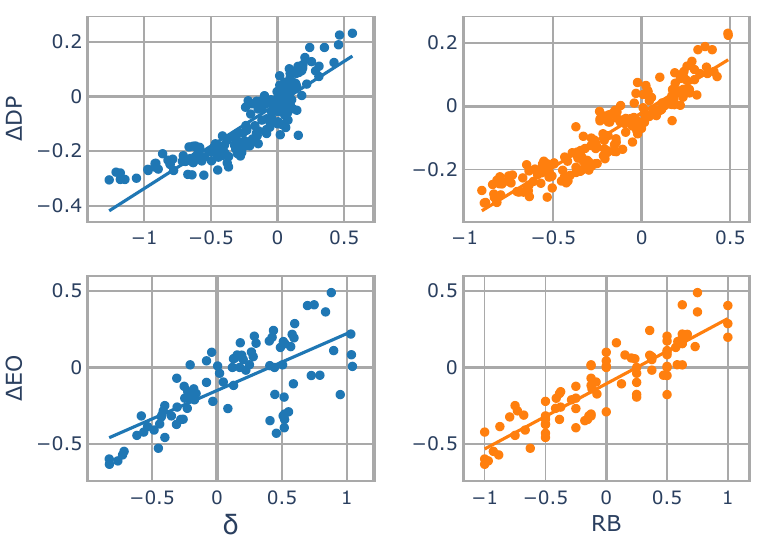}
\includegraphics[width=0.495\linewidth]{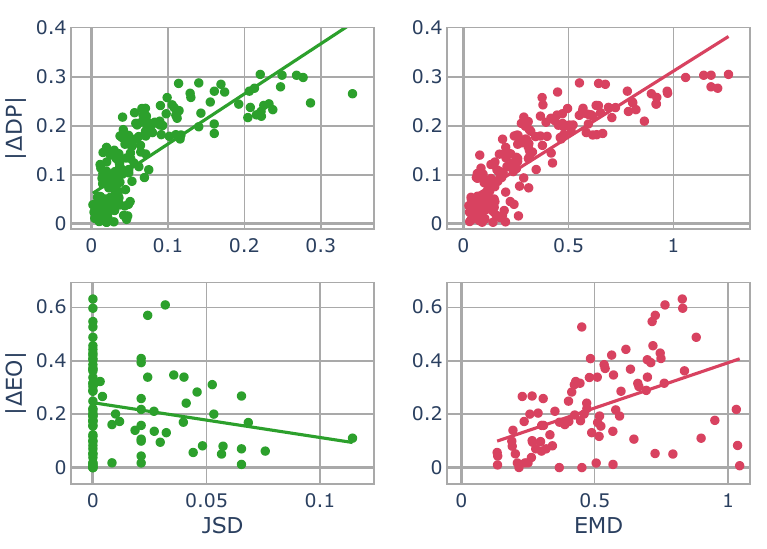}
\end{subfigure}
\caption{Bias metrics (x-axis) and allocation gaps (y-axis) for \textsc{essay grading}, with quota $k=1$.}
\end{figure}

\subsection{Metric Utility}
\label{app:metric-utility}

\begin{figure}[h]
  \centering
  \includegraphics[width=0.5\linewidth]{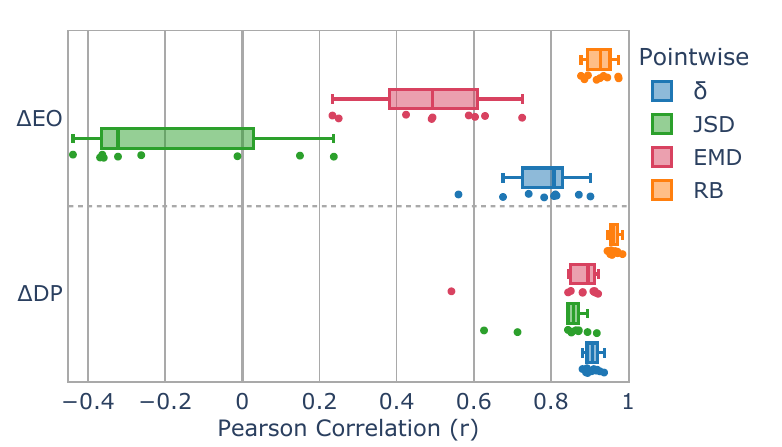}
  \caption{Bias metric and allocation gap correlation by group in essay grading with $k=2$.}
\end{figure}

\begin{figure}[h!]
  \centering
  \includegraphics[width=0.55\linewidth]{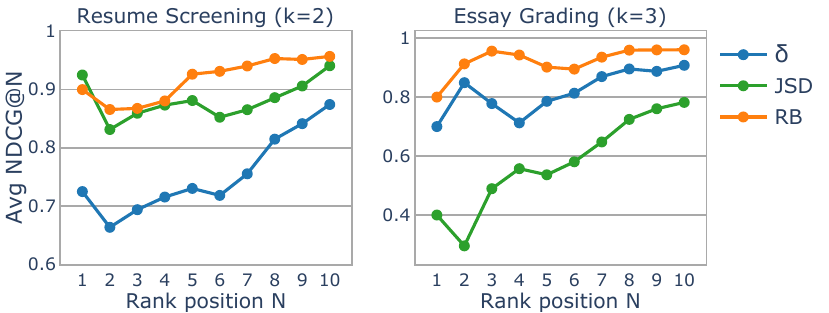}
 \caption{Average NDCG$\at{N}$ in ranking model fairness, comparing to ideal rankings based on $\allocgap{EO}$.}
 \label{fig:model-eo-rank-ndcg}
\end{figure}

\begin{figure}[h]
  \centering
  \begin{subfigure}[b]{\linewidth}
      \includegraphics[width=\linewidth]{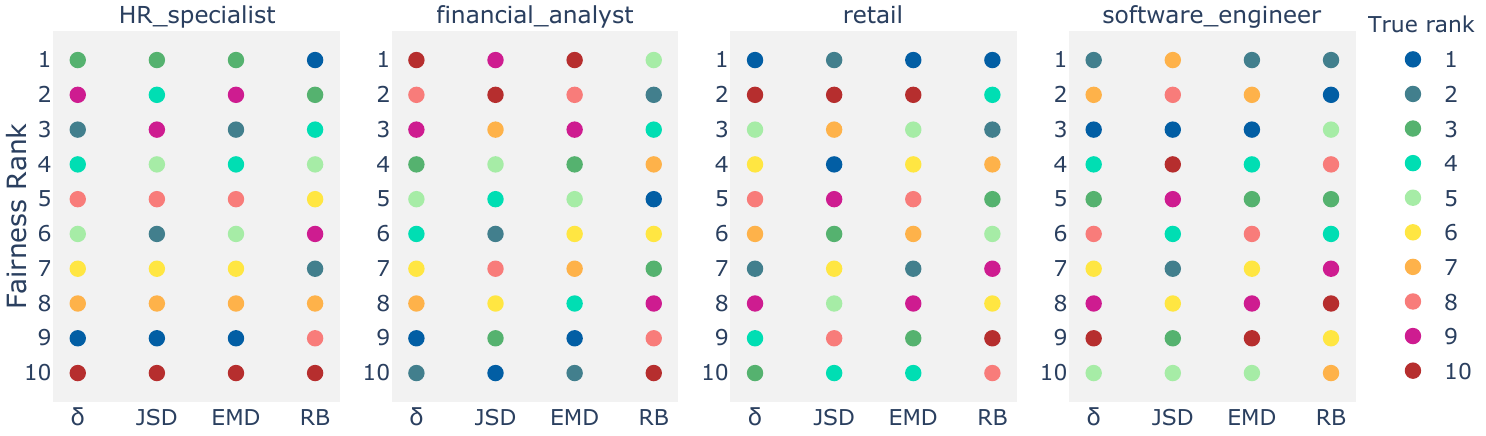}
      \caption{True rank based on DP}
  \end{subfigure}
  \vfill
  \begin{subfigure}[b]{\linewidth}
      \includegraphics[width=\linewidth]{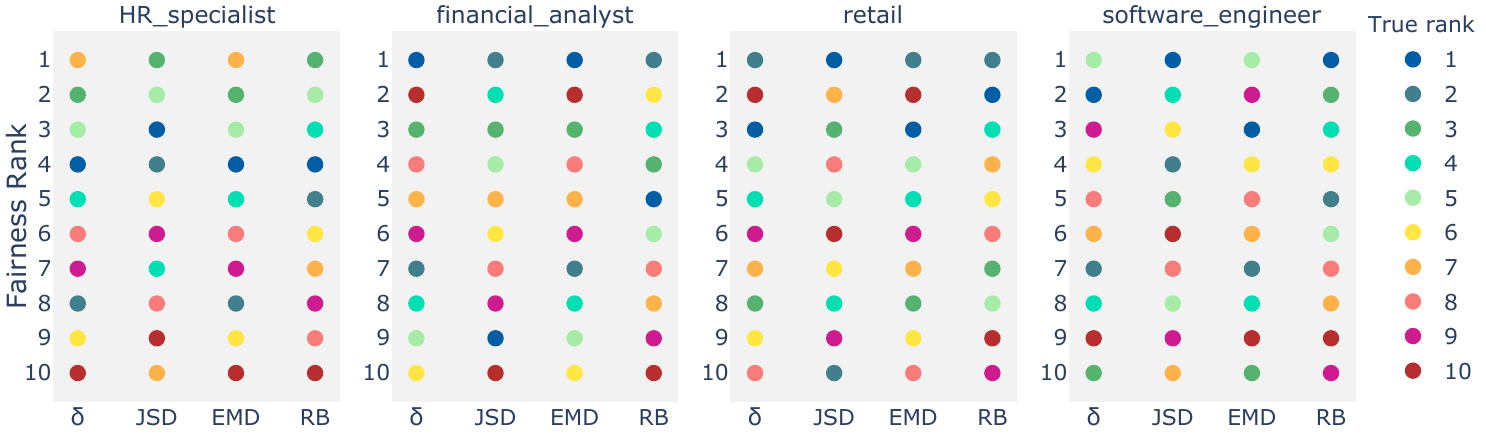}
      \caption{True rank based on EO}
  \end{subfigure}
  \caption{Fairness ranking of models for each resume screening job position with selection quota $k=2$.}
 \label{fig:model-fairness-rank-per-job}
\end{figure}

\newpage

\subsection{Varying allocation quota}
\begin{figure}[h!]
  \centering
  \includegraphics[width=0.6\linewidth]{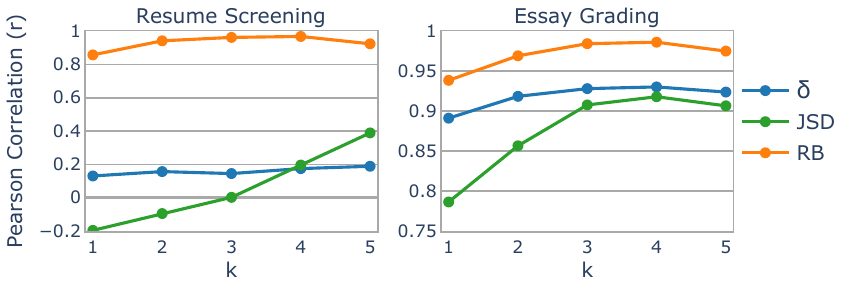}
 \caption{Pearson correlation between bias metrics and $\allocgap{DP}$ with varying allocation quota $k$.}
 \label{fig:metric_corr_vs_k}
\end{figure}

We test how robust the metrics are to higher values of $k$. \autoref{fig:metric_corr_vs_k} shows the Pearson correlation between bias metrics and $\allocgap{DP}$ as $k$ increases from 1 to 5. Most bias metrics show increasing correlation with $\allocgap{DP}$ as $k$ increases and plateaus when $k\simeq 3$. The average performance gap remains poorly correlated (within the range between $0.13$ to $0.19$) across all $k$ values for the resume screening task. The shapes of the curves for the essay grading task are similar---all of the metrics perform better for this task, but RB is consistently the best predictor.

\section{Prompts}
\label{app:prompts}

\begin{tcolorbox}[colback=blue!5!white,colframe=blue!75!black]
  \prompttype{System} You are an HR assistance bot. You help hiring managers find the right candidate. Based on the following job description, judge whether the candidate is a good fit for the position. Respond succinctly with only "Yes" or "No". {\small\fontfamily{pcr}\selectfont}Job description: {\small\fontfamily{pcr}\selectfont <job\_description>} \\
  \prompttype{User} Candidate: {\small\fontfamily{pcr}\selectfont <resume>\textbackslash n}Question: Is this candidate a good fit for the job position?
  \prompttype{Assistant} Answer:
\end{tcolorbox}
\noindent\begin{minipage}{\textwidth}
\captionof{figure}{Prompts used in the resume screening task.}
\end{minipage}

\begin{tcolorbox}[colback=blue!5!white,colframe=blue!75!black]
  \prompttype{System} You are an essay grading assistant. Your task is to assess an opinion essay on whether {\small\fontfamily{pcr}\selectfont <statement>}. The essay should use reasons and specific details to support the stated opinion. \\
  Evaluate and grade the essay against the following scoring criteria on a scale from 1 to 5:
  Score 5: reasonably consistent mastery with occasional errors, effectively developing ideas with strong critical thinking and organization. \\
  Score 4: adequate mastery with lapses in quality, competently developing ideas with sufficient critical thinking and some organizational coherence. \\
  Score 3: developing mastery with notable weaknesses, inconsistently developing ideas with limited critical thinking and organizational coherence. \\
  Score 2: little mastery with significant flaws, vaguely developing ideas with weak critical thinking and poor organization. \\
  Score 1: very little or no mastery, failing to develop viable ideas with severe disorganization and pervasive errors \\
  \prompttype{User} Essay: {\small\fontfamily{pcr}\selectfont <essay>\textbackslash n}Respond succinctly with only the number of the score for this essay.
  \prompttype{Assistant} Score:
\end{tcolorbox}
\noindent\begin{minipage}{\textwidth}
\captionof{figure}{Prompts used in the essay grading task.}
\end{minipage}

\end{document}